\documentclass[journal]{IEEEtran}
\usepackage{amsmath,amsfonts}
\usepackage{mathtools}
\usepackage{algorithmic}
\usepackage{algorithm}
\usepackage{array}
\usepackage[caption=false,font=normalsize,labelfont=sf,textfont=sf]{subfig}
\usepackage{textcomp}
\usepackage{stfloats}
\usepackage{url}
\usepackage{verbatim}
\usepackage{graphicx}
\usepackage{cite}
\usepackage{dirtytalk}
\usepackage{booktabs}
\usepackage{float}
\usepackage{multicol}
\usepackage{color}
\usepackage{pifont}
\usepackage{colortbl}
\usepackage{multirow}
\usepackage{pifont}
\usepackage[table]{xcolor}
\begin{document}
\bstctlcite{IEEEexample:BSTcontrol}

\title{A Signal Contract for Online Language Grounding and Discovery in Decision-Making}

% \title{Grounding and Discovery Services for Language-Assisted Decision-Making: A Signal Contract Architecture}
% LUCIFER: Middleware for Language-Driven Assistance via a Signal Contract
%A Client-Agnostic Signal Contract for Language-to-Decision Adaptation

%Runtime Language-to-Control Signals via Middleware Contracts
% Signal Contracts for Runtime Autonomy Assistance

% Middleware Signal Contracts for Human-in-the-Loop Autonomy

\author{Dimitris Panagopoulos$^{1}$, Adolfo Perrusqu\'{i}a$^{1}$ and Weisi Guo$^{1}$% <-this % stops a space
\thanks{$^{1}$Faculty of Engineering and Applied Science, Cranfield University, Cranfield MK43 0AL, UK,
        {\tt\small \{d.panagopoulos, adolfo.perrusquia-guzman, weisi.guo\}@cranfield.ac.uk}. 
        This work is funded by EPSRC iCASE with Thales UK (EP/X52475X/1)}%
}

\markboth{Submitted to the IEEE for possible publication. Copyright may be transferred without notice.}%
{Shell \MakeLowercase{\textit{et al.}}: A Sample Article Using IEEEtran.cls for IEEE Journals}

% \markboth{IEEE Transactions on Systems, Man, and Cybernetics: Systems,~Vol.~XX, No.~XX, April~2025}%
% {Panagopoulos \MakeLowercase{\textit{et al.}}: Language Understanding and Context-Infused Framework for Exploration and Behavior Refinement}

%\IEEEpubid{00--00/00\$0.0~\copyright~2021 IEEE}

\maketitle
\begin{abstract}
Autonomous systems increasingly receive time-sensitive contextual updates from humans through natural language, yet embedding language understanding inside decision-makers couples grounding to learning or planning. This increases redeployment burden when language conventions or domain knowledge change and can hinder diagnosability by confounding grounding errors with control errors. We address \emph{online language grounding} where messy, evolving verbal reports are converted into control-relevant signals during execution through an interface that localises language updates while keeping downstream decision-makers language-agnostic. We propose \textsc{LUCIFER} (Language Understanding and Context-Infused Framework for Exploration and Behavior Refinement), an inference-only middleware that exposes a \emph{Signal Contract}. The contract provides four outputs, policy priors, reward potentials, admissible-option constraints, and telemetry-based action prediction for efficient information gathering. We validate \textsc{LUCIFER} in a search-and-rescue (SAR)-inspired testbed using dual-phase, dual-client evaluation: (i) component benchmarks show reasoning-based extraction remains robust on self-correcting reports where pattern-matching baselines degrade, and (ii) system-level ablations with two structurally distinct clients (hierarchical RL and a hybrid A$^\star$+heuristics planner) show consistent necessity and synergy. Grounding improves safety, discovery improves information-collection efficiency, and only their combination achieves both.
\end{abstract}


% The paper should hammer home:

% 1. **Middleware contribution**: Training-decoupled signal contract (Ψ, ΦΨ, A', a*)
% 2. **Pattern consistency**: Grounding→safety, discovery→efficiency, synergy—**holds for both clients**
% 3. **Client-agnosticism validated**: RL learns from signals over episodes, hybrid executes signals immediately—**same pattern emerges**
% 4. **Performance gap is operational**: Planning with perfect knowledge achieves ceiling, learning under sparse feedback has sample complexity cost—**this is expected and interpretable**
% 5. **Not a paradigm comparison**: We don't claim RL>planning or vice versa; we claim **the middleware works regardless of client type**

\section{Introduction}

\subsection{The Online Language-Grounding Problem}

Autonomous systems operating in high-stakes environments increasingly depend on time-sensitive updates from human stakeholders~\cite{saeed2019role,kruijff2013experience}.
These declarative updates (e.g., safety reports, operator instructions)~\cite{da2024survey} arrive as unstructured natural language, yet most autonomy stacks lack a principled way to ground them into control-relevant quantities during execution~\cite{tellex2020robots}. This creates an \emph{asymmetry} where meaning (semantics) of information is expressed in human language, while the final, actionable decisions are made based on numerical, mathematical representations.

A common response is to place language understanding \emph{inside} the learner or planner (e.g., language-conditioned policies or controllers). However, embedding grounding within the decision-maker couples (i) language conventions, (ii) domain knowledge, and (iii) optimisation/training dynamics, raising redeployment burden when language or constraints change and reducing diagnosability by confounding grounding errors with control errors~\cite{cohen2024survey,karamcheti2022lila}.

\textbf{Architectural positioning.}
We therefore study deployment-time declarative reports that modify what is safe, feasible, or desirable, and we adopt an externalised grounding architecture. In particular, language processing runs in middleware, while downstream decision-makers remain language-agnostic and consume only standardised numerical signals through a stable interface. This separation allows us to localise language updates, improve failure isolation, while supporting heterogeneous clients without embedding language inside their optimisation loops.

In this setting, two runtime requirements arise. \textbf{Grounding} must convert messy, evolving reports into enforceable decision signals that improve safety, and \textbf{Discovery} must efficiently select high-value information-gathering actions in large query spaces so that the system does not waste interaction budget on trial-and-error. Urban search and rescue (USAR) exemplifies this regime where robots can perceive structure, but first-responder and survivor reports often carry the most actionable semantics that may be incomplete or self-correcting~\cite{murphy2000mobility,kruijff2013experience,saeed2019role,ter2023stakeholder,pirinen2022aerial,gruffeille2024disaster}. We use USAR as the motivating domain and evaluate in a USAR-inspired simulation testbed.

% Taxonomies distinguishing \emph{language-conditioned} from \emph{language-assisted} settings~\cite{luketina2019surveyreinforcementlearninginformed,cao2024survey}
% help position our problem. We study deployment-time reports that modify what is safe, feasible, or desirable.
% Our contribution is infrastructure that standardises how such reports become client-consumable decision signals.

\subsection{\textsc{LUCIFER}: Middleware for Online Adaptation}
We introduce \textbf{LUCIFER} (Language Understanding and Context-Infused Framework for Exploration and Behavior Refinement), a \textit{training-decoupled middleware} that implements this externalised grounding architecture and exposes a client-agnostic Signal Contract. Grounding maps streaming language to standardised signals (policy priors, reward potentials, and admissible-option constraints), while a discovery service predicts high-value information-gathering actions from client-agnostic telemetry (trace summaries). We ask: \emph{can online language processing be externalised from a decision-maker via a stable Signal Contract, while still yielding safe and efficient behaviour during execution?} Our claim is sufficiency of an inference-only, externalised grounding and discovery layer in the validated setting and not superiority over coupled, end-to-end language-conditioned systems. Accordingly, we validate LUCIFER through (i) component benchmarks and (ii) system-level necessity/synergy ablations under fixed downstream clients.

\textbf{Signal Contract (high-level).}
LUCIFER communicates with downstream agents through a minimal \emph{Signal Contract} that externalises language into a small set of consumable decision signals. Concretely, the middleware emits (i) priors that bias action selection, (ii) potentials that shape exploration, (iii) constraints that mask unsafe or infeasible options, and (iv) action predictions for targeted information gathering. The contract is intentionally client-agnostic meaning that any learner or planner that can consume these signals can benefit without retraining its core optimisation on language. This enables rapid updates when reports evolve and improves diagnosability by separating grounding failures from control failures.

\begin{figure*}[t]
    \centering
    \includegraphics[width=0.6\textwidth]{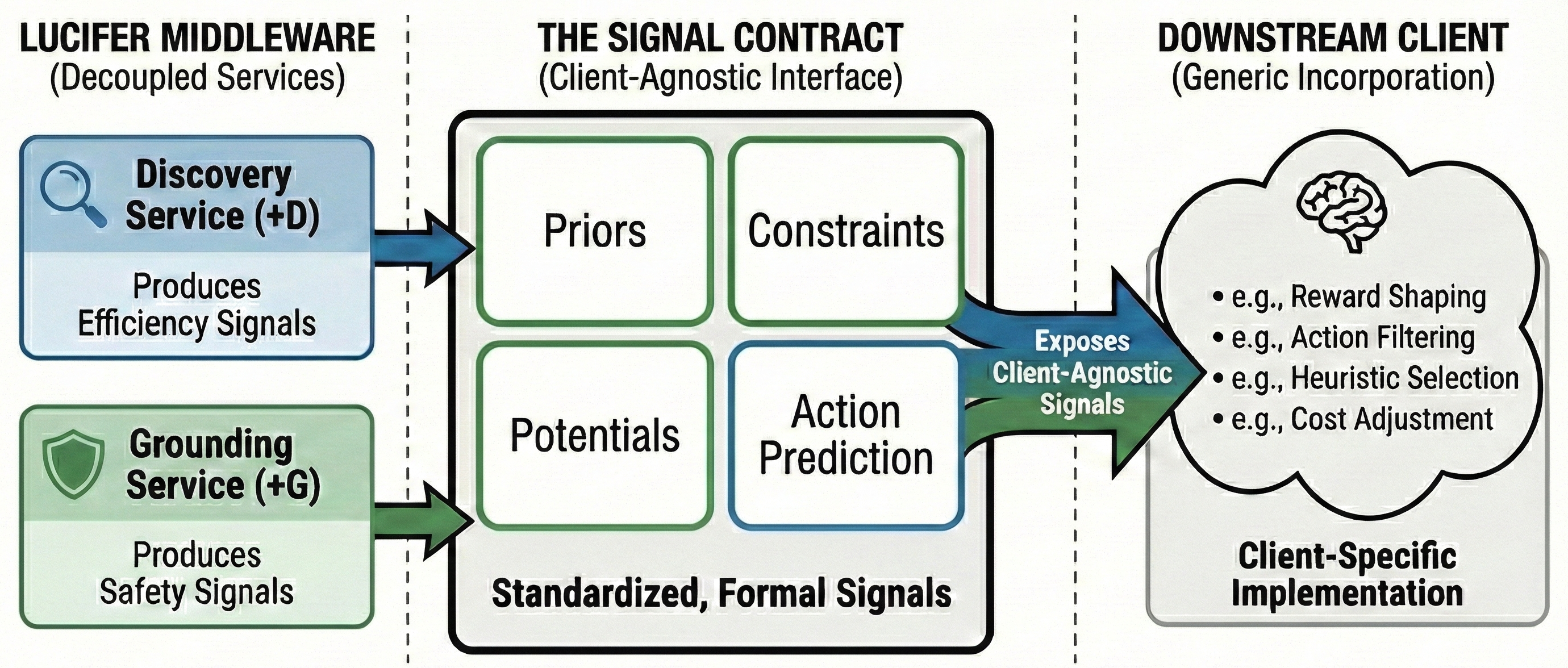}
    \caption{\textbf{The \textsc{LUCIFER} Architecture.} Middleware services independently produce safety- and efficiency-relevant outputs and expose them only through a \textit{Signal Contract}:
policy priors, reward potentials, admissible-action constraints, and action prediction.
Downstream clients consume these abstract signals using native mechanisms (e.g., action filtering, reward shaping), remaining language-agnostic and decoupled from middleware internals.}
    \label{fig1:lucifer_framework}
\end{figure*}

\subsection{Contributions}

We make the following contributions:
\begin{enumerate}
    \item \textbf{Training-decoupled middleware for online grounding and discovery.}
    We introduce \textsc{LUCIFER}, an inference-only middleware layer that externalises online language grounding and discovery,
    translating streaming language and client-agnostic telemetry into control-relevant signals without reading or depending on client-internal optimisation state.

    \item \textbf{A Signal Contract with four client-agnostic outputs.}
    We formalise a standard contract exposing \emph{priors}, \emph{potentials}, \emph{constraints}, and \emph{action predictions} as consumable numerical signals that enable safety-aware adaptation.

    \item \textbf{Discovery as telemetry-only action prediction.}
    We provide a discovery service that recommends high-value query actions from trace summaries, without accessing client parameters, gradients, or value estimates. This improves information-collection efficiency by replacing trial-and-error query selection with informed predictions, addressing the sample inefficiency of grounding-only configurations.

    \item \textbf{Necessity and synergy via dual-phase, dual-client evaluation.}
    We validate (i) component robustness, where grounding maintains 91--100\% accuracy on messy, self-correcting language while pattern-matching baselines collapse to 20--36\%, and (ii) system-level ablations, where grounding-only improves safety without efficiency, discovery-only improves efficiency without safety, and the combined middleware achieves both. This pattern holds consistently for learning (RL) and non-learning (hybrid planner) clients.
\end{enumerate}

\section{Related Work}
\label{sec:related}
We organise prior work along two practical axes: \emph{when} language is integrated (offline vs.\ online) and \emph{where} language understanding resides (coupled inside the learner/planner loop vs.\ externalised as a module). We use these axes to situate our approach relative to instruction-following control, design-time language assistance, and language-in-the-loop learning.

\subsection{Static Instructions and Modular Planning: Language-Conditioned Control}
Language-conditioned RL and planning typically treat language as clean instructions that remain fixed over an episode (e.g., ``go to the red room'')~\cite{luketina2019surveyreinforcementlearninginformed,liu2022instruction}.
Recent robotic systems leverage vision--language models (VLMs) to score candidate navigation plans against such static directives~\cite{shah2023navigation,lynch2023interactive}.
Similarly, LLM-based planners such as PaLM-SayCan ground an initial command into a sequence of executable skills by combining language priors with affordance/value estimates~\cite{ahn2022can}.
Other modular grounding systems such as SayPlan use structured scene representations (e.g., 3D scene graphs) to ground LLM-generated plans in large environments, often with iterative replanning and classical planning components~\cite{rana2023sayplan}.

These approaches are effective for instruction following and plan generation, but they are primarily imperative (goal/task specification) and typically assume language is given upfront, not discovered as evolving constraints. In contrast, we target declarative reports that modify feasibility and learning signals during execution, with grounding performed externally and exposed to clients only as decision signals.

\subsection{Offline Assistance: Language for Reward Design and Exploration (Design-Time)}
A complementary paradigm uses language to assist RL via \emph{offline} injection of prior knowledge.
Frameworks such as Text2Reward~\cite{xie2023text2reward} and Eureka~\cite{ma2023eureka} use LLMs to synthesise executable reward code from task descriptions, enabling reward engineering without manual programming.
Others use LLMs to propose intrinsic goals or curricula to guide exploration (e.g., ELLM and autotelic agents with language-generated goals)~\cite{du2023guiding,colas2023augmenting,song2023llm}.
Relatedly, pretrained vision--language models can serve as \emph{zero-shot reward models} from natural-language prompts (VLM-RM), providing a numeric reward signal that enables training without handcrafted rewards~\cite{rocamonde2023vision}.

While these methods strongly leverage language, the integration is primarily before training with language specifying objectives, rewards, or curricula. However, deployed policies do not generally consume streaming human reports and therefore do not update constraints online without additional mechanisms. By comparison, we translate \emph{deployment-time} reports into multiple signal types (bias, shaping, constraints, and query guidance) without coupling language understanding to client optimisation.

\subsection{Online Feedback and Interfaces: Language in the Loop}
Recent work moves toward online integration, often by placing language feedback \emph{inside} the learning interface.
LLF-Bench formalizes interactive learning from language feedback, replacing scalar rewards with textual feedback and instructions~\cite{cheng2023llfbenchbenchmarkinteractivelearning}.
In shared autonomy, LILA learns language-informed latent control interfaces that allow users to influence robot behavior via natural language during execution~\cite{karamcheti2022lila}. Similarly, ECLAIR~\cite{tarakli2024interactive} uses LLMs to interpret diverse natural language feedback (evaluative, corrective, and guidance) directly within an interactive RL framework, enabling human teachers to shape robot behaviour through speech.
Safety-oriented systems also incorporate online language and structured knowledge. For example, integrating LLM prompting with embodied knowledge graphs to validate or constrain robot actions toward safer behavior~\cite{qi2024safety}.

These approaches provide important mechanisms for online correction, preference shaping, or action validation, but they commonly couple language processing to the agent/controller interface (e.g., language feedback becomes part of the training signal, or language modulates a learned control interface). By comparison, we keep the downstream client language-agnostic by translating reports into a numerical representation interface, exposing consumable signals rather than embedding language feedback inside the client's decision-making.

\subsection{Situating LUCIFER Within Existing Taxonomies}
Surveys of language integration in sequential decision-making distinguish settings where language is a policy input from those where language provides auxiliary guidance~\cite{luketina2019surveyreinforcementlearninginformed}. These categories are MDP-level distinctions describing how language enters the learning system (as an observation/policy input or as part of the objective/design).

LUCIFER is architecturally different. It is \textit{middleware} that performs language processing externally and exposes only signals to a downstream client. The key distinction is not a new taxonomy category, but \emph{where processing happens} and \emph{whether it is coupled to optimisation}. Prior systems either (i) embed language into policy/planning interfaces for instruction following~\cite{ahn2022can,rana2023sayplan}, or (ii) use language to define rewards/objectives at design time~\cite{xie2023text2reward,ma2023eureka,rocamonde2023vision}, or (iii) place language feedback inside the learning loop~\cite{cheng2023llfbenchbenchmarkinteractivelearning}. LUCIFER instead enforces a training-decoupled grounding boundary where language is processed via inference-only middleware, and the client receives only numerical quantities through a stable interface regardless of that client being a learning agent, a planner, or a rule-based controller.

We empirically validate this architectural distinction by testing the same interface with two structurally different downstream clients (a learning-based agent and a non-learning planner), holding each client fixed across ablations.

\section{Methodology}\label{section:methodology}

\subsection{Architectural Overview}
\textsc{LUCIFER} (Fig.~\ref{fig1:lucifer_framework}) is a middleware layer that sits between streaming human reports and downstream decision-makers. Clients provide only minimal inputs (reports and lightweight telemetry), and the middleware returns a fixed set of control-relevant signals via the \emph{Signal Contract}. Clients remain responsible for how these signals are incorporated.

\textbf{Notation and interfaces.}
Let $\mathcal{V}$ denote the space of verbal reports, $\mathcal{I}$ the Information Space (semantic categories), and $\mathcal{B}$ a domain knowledge base used for grounding.
Downstream clients act over a client-defined decision-context space $\mathcal{X}$ (e.g., grid locations or compact location signatures). At context $x\in\mathcal{X}$, the available option set is $U(x)$, and the global option alphabet is $U=\bigcup_{x\in\mathcal{X}}U(x)$.
Grounding produces structured semantic objects
$
C \subseteq \mathcal{E}\times\mathcal{I}\times\mathcal{X},
$
where $\mathcal{E}$ is the set of extracted entities.
Discovery consumes client-agnostic telemetry summaries: an episodic trace summary $\xi\in\Xi$ and a cross-episode telemetry memory $\mathcal{D}\in\Delta$ (a log of trace--outcome records).

\textbf{Middleware services.} \textsc{LUCIFER} comprises two architecturally independent services:
\begin{enumerate}
    \item \textbf{Grounding (language $\rightarrow$ signals):}
    configured by the Information Space $\mathcal{I}$ and knowledge base $\mathcal{B}$.
    A Context Extractor $\mathcal{E}_C$ translates and maps a report $v \in \mathcal{V}$ into structured
    semantic objects $C \subseteq \mathcal{E}\times\mathcal{I}\times\mathcal{X}$, which the middleware
    transduces into contract signals.
    \item \textbf{Discovery (telemetry $\rightarrow$ prediction):}
    an Exploration Facilitator $\mathcal{E}_F$ that consumes only client-agnostic telemetry
    (state/action/reward/event traces summarized as $(x_t,\xi,\mathcal{D})$) and outputs an advisory option prediction $u^\star\in U(x_t)$
    at designated information-gathering decision points.
\end{enumerate}

\textbf{Signal Contract outputs.}
\begin{itemize}
    \item \textbf{Grounding signals:} policy priors $\Psi_x(u)\in\mathbb{R}$, reward potentials $\Phi_{\Psi}(x)\in\mathbb{R}$, and admissible-option constraints $U'(x)\subseteq U(x)$.
    \item \textbf{Discovery signal:} an option prediction $u^\star\in U(x)$ produced from telemetry only.
\end{itemize}
In practice, these signals enter the client as additive biases (priors), shaped rewards/heuristics (potentials), option masking (constraints), and query-action suggestions (prediction).

\textbf{Downstream clients.} We evaluate the contract with two structurally distinct consumers:
(i) a learning-based hierarchical RL client and (ii) a non-learning hybrid planner.
Both expose only minimal telemetry and consume the same contract outputs with neither processing raw text.
We hold each client fixed across ablations and vary only which contract signals are enabled, attributing
effects to the middleware pattern rather than to client tuning (Sec.~\ref{sec:experimental_setup}).

\subsection{Design Rationale: Architectural Decisions and Trade-offs}
We motivate three design choices: (i) why the contract exposes four outputs,
(ii) why the middleware is training-decoupled, and (iii) why grounding and discovery
remain separate services.

\paragraph{Why four signal types?}
The contract spans four complementary adaptation pathways:
\emph{policy bias} (priors), \emph{reward guidance} (potentials), \emph{feasibility enforcement}
(constraints), and \emph{query efficiency} (action prediction). Each addresses a distinct runtime
failure mode. Priors provide immediate directional bias, potentials provide longer-horizon shaped feedback, constraints enforce hard safety boundaries, and action recommendation reduces wasted queries by exploiting cross-episodic telemetry patterns. Separating these pathways avoids conflating soft guidance with hard feasibility and allows partial activation when only some information is available or trusted.

\paragraph{Why enforce training decoupling?}
The middleware does not access client parameters, gradients, or internal value estimates. This constraint (i) localises updates to middleware configuration (e.g., $\mathcal{I},\mathcal{B}$), (ii) permits heterogeneous consumers of the same interface, and (iii) isolates grounding faults from control/learning faults for direct debugging and replacement. We evaluate this boundary for sufficiency in our setting rather than claiming universal superiority over coupled training.

\paragraph{Why separate grounding from discovery?}
Grounding and discovery consume different inputs and fail differently. In particular, grounding translates reports into
control-relevant safety/utility signals, while discovery predicts high-value query actions from telemetry.
Keeping them separate enables independent benchmarking and graceful degradation when only one input stream (reports or telemetry) is reliable.

% pseudocode for LLM-1-ContextExtractor
\begin{algorithm}[t]
\caption{Context Extractor Service ($\mathcal{E}_C$)}
\label{alg:context_extractor}
\begin{algorithmic}[1]
\REQUIRE Verbal Input Stream $v_{1:k}$, Knowledge Base $\mathcal{B}$, Information Space $\mathcal{I}$
\ENSURE Structured Semantic Objects $C$

\STATE \textbf{// Service configuration}
\STATE \textit{Initialize RAG-augmented LLM with retrieval over $\mathcal{B}$}
\STATE $C \leftarrow \emptyset$

\STATE \textbf{// Stream Processing}
\FORALL{verbal report $v_i \in v_{1:k}$}
    \STATE \textit{Extract candidate entities:}
    \STATE $E_i \leftarrow \textsc{ExtractEntities}(v_i,\mathcal{B})$
    
    \FORALL{entity $e_j \in E_i$}
        \STATE \textbf{// Schema Mapping}
        \STATE \textit{Map entity to Information Space category:}
        \STATE $\mathrm{cat}_j \leftarrow \textsc{Classify}(e_j,\mathcal{I})$
        
        \STATE \textit{Ground to a client-defined decision context:}
        \STATE $x_j \leftarrow \textsc{GroundToContext}(e_j,\mathcal{B})$
        
        \STATE \textit{Construct structured object:}
        \STATE $c_j \leftarrow (e_j,\mathrm{cat}_j,x_j)$
        \STATE $C \leftarrow C \cup \{c_j\}$
    \ENDFOR
\ENDFOR

\RETURN $C$ \textit{// Consumed by the Signal Contract transducer}
\end{algorithmic}
\end{algorithm}

\begin{algorithm}[ht]
\caption{Exploration Facilitator Service ($\mathcal{E}_F$)}
\label{alg:exploration_facilitator}
\begin{algorithmic}[1]
\REQUIRE Telemetry request $(x_t,\xi,\mathcal{D})$ with $\xi\in\Xi$, $\mathcal{D}\in\Delta$, candidate option set $U(x_t)$
\ENSURE Recommended option $u^\star\in U(x_t)$
\STATE \textbf{// Middleware Internal Encoding}
\STATE $prompt \leftarrow \textsc{EncodeToText}(x_t,\xi,\mathcal{D})$
\STATE \textbf{// Zero-Shot Reasoning Service}
\STATE $P(u \mid x_t,\xi,\mathcal{D}) \leftarrow \mathcal{LLM}(prompt), \quad u\in U(x_t)$
\STATE \textbf{// Signal Generation}
\STATE $u^\star \leftarrow \arg\max_{u \in U(x_t)} P(u \mid x_t,\xi,\mathcal{D})$
\RETURN $u^\star$
\end{algorithmic}
\end{algorithm}

\subsection{Information Space}\label{subsec:Information_Space}
The Information Space $\mathcal{I}=\{I_1,\dots,I_m\}$ is a structured set of mission-relevant
categories (e.g., hazard, safe zone, victim). It serves as the grounding schema used to map
linguistic entities to operational categories, independent of the downstream client's control
paradigm.

\subsection{Context Extractor: LLM as Semantic Parser}\label{subsec:Context_Extractor}
The Context Extractor ($\mathcal{E}_C$) is invoked as a middleware service at runtime and converts
streaming verbal reports into structured semantic objects consumable by the Signal Contract.
Given a report $v\in\mathcal{V}$, it outputs
\[
C=\{c_1,\dots,c_n\}, \qquad c_j=(e_j,\mathrm{cat}_j,x_j),
\]
where each object contains an entity $e_j\in\mathcal{E}$, its Information Space category $\mathrm{cat}_j\in\mathcal{I}$, and a grounded decision context $x_j\in\mathcal{X}$.

The component uses a large language model (LLM) augmented with Retrieval-Augmented Generation (RAG)~\cite{lewis2020retrieval}
over a domain knowledge base $\mathcal{B}$ to resolve ambiguity (e.g., ``the restaurant'' $\rightarrow$ a unique context in $\mathcal{X}$),
interpret terminology (e.g., ``rubble zone'' $\rightarrow$ hazard), and handle self-corrections and implied references.
These robustness regimes are used as stress tests for known phenomena (disfluency/self-repair and implicit reference resolution)~\cite{1571698600491774336,hou-etal-2018-unrestricted}. We do not propose a new linguistic taxonomy.

Formally, $\mathcal{E}_C:\mathcal{V}\times\mathcal{B}\rightarrow 2^{\mathcal{E}\times\mathcal{I}\times\mathcal{X}}$.
Algorithm~\ref{alg:context_extractor} details the extraction process. The extractor operates online: upon receiving
a report, the client synchronously invokes $\mathcal{E}_C$ and receives $C$ for immediate contract-level signal generation.

\subsection{Exploration Facilitator: LLM as Zero-Shot Predictor}\label{subsec:Exploration_Facilitator}
\begin{figure}[t]
\centering
\includegraphics[width=\columnwidth]{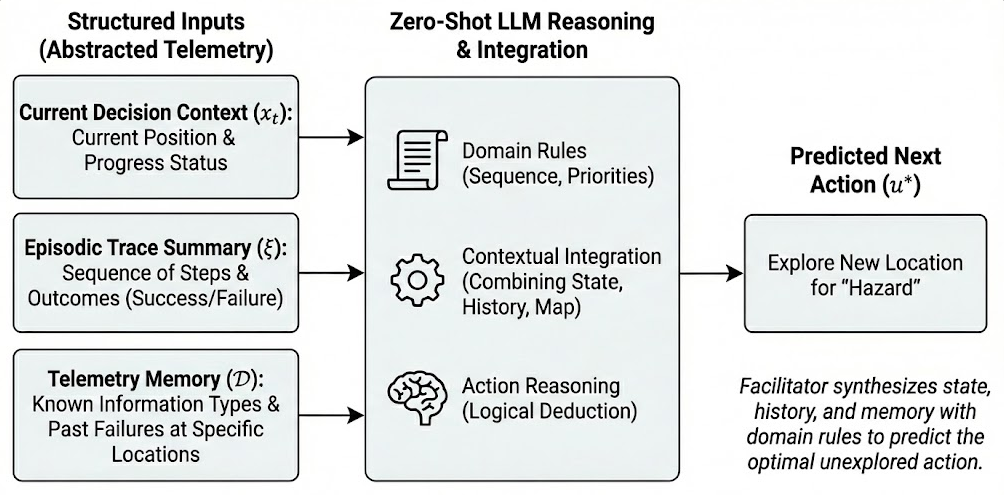}
\caption{\textit{Exploration Facilitator (telemetry-based discovery).} The facilitator constructs a prompt from client-agnostic telemetry: current decision context ($x_t$), episodic trace ($\xi$), and cross-episodic telemetry memory ($\mathcal{D}$). An LLM performs zero-shot reasoning to propose an advisory query option ($u^\star$) likely to yield high-value information.}
\label{fig2:facilitator}
\end{figure}

The Exploration Facilitator ($\mathcal{E}_F$) provides telemetry-only guidance to accelerate information discovery (Fig.~\ref{fig2:facilitator}). It consumes trace summaries of recent decisions and outcomes and returns an advisory query-action recommendation that the client may accept or ignore.

At designated information-gathering decision points, the service constructs a prompt from:
\begin{enumerate}
    \item \textbf{Current decision context ($x_t$):} a compact context token (e.g., a location signature).
    \item \textbf{Episodic trace summary ($\xi$):} a short window of recent query attempts and outcomes.
    \item \textbf{Telemetry memory ($\mathcal{D}$):} cross-episode records linking trace patterns to outcomes.
\end{enumerate}
The LLM returns a distribution over available query options $u \in U(x_t)$:
\[
P(u \mid x_t,\xi,\mathcal{D}),
\qquad
u^\star=\arg\max_{u \in U(x_t)} P(u \mid x_t,\xi,\mathcal{D}).
\]
Formally, $\mathcal{E}_F:\mathcal{X}\times\Xi\times\Delta\rightarrow U$ with output $u^\star\in U(x_t)$.
Algorithm~\ref{alg:exploration_facilitator} specifies the service. This signal enables clients to reduce trial-and-error
in large option spaces without exposing policy parameters, value estimates, or gradients.

% figure

% figure 
\begin{figure}[t]
\centering
\includegraphics[width=\columnwidth]{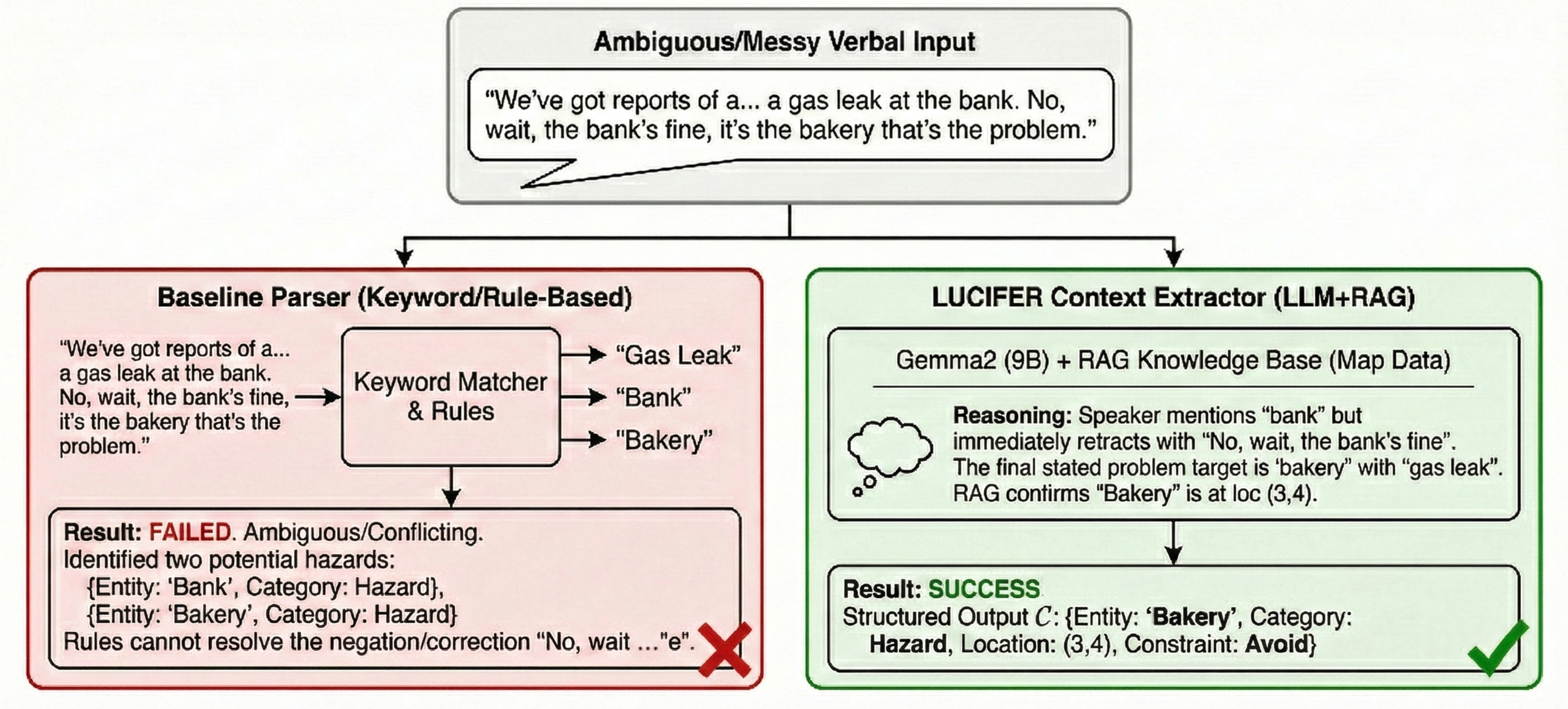}
\caption{\textit{Robustness comparison.} Baseline parsers fail on ambiguous inputs with self-corrections (left), incorrectly treating retracted entities (e.g., the bank) as valid. In contrast, LUCIFER's grounding service (right) uses semantic reasoning to isolate the true target (the bakery) and produce correct grounded constraints.}
\label{fig3:verbal}
\end{figure}

\subsection{The Signal Contract: A Formal Client-Agnostic Interface}\label{subsec:Signal_Contract}
The Signal Contract is the interface between middleware and client, transducing grounded semantics and telemetry-derived insights into four standardised signals consumable by heterogeneous decision-makers. Let $x\in\mathcal{X}$ be the current decision context with available options $U(x)$.

\textbf{Signal 1: Policy Priors $\Psi_x(u)$.}
A scoring function over available options, $\Psi:\mathcal{X}\times U \rightarrow \mathbb{R}$, encoding
immediate directional preferences. Priors are derived from grounded semantics (e.g., regions classified as
undesirable/desirable/critical) and enable immediate biasing of decision selection (e.g., discouraging options
leading toward hazards) without requiring any particular optimisation method.

\textbf{Signal 2: Reward Potentials $\Phi_{\Psi}(x)$.}
A scalar potential $\Phi_{\Psi}:\mathcal{X}\rightarrow\mathbb{R}$ that makes contexts semantically attractive
or repulsive. This signal supports potential-based shaping when a client uses reward-driven optimisation, while
remaining a general utility signal for other paradigms (e.g., heuristic augmentation).

\textbf{Signal 3: Admissible-Option Constraints $U'(x) \subseteq U(x)$.}
A hard feasibility filter that prunes (or restricts) the option set based on grounded constraints.
Let $X_u,X_d,X_o \subseteq \mathcal{X}$ denote undesirable, desirable, and critical context sets inferred from grounded reports.
Using a generic one-step outcome operator $f(x,u)$ (e.g., successor generation in planning or environment stepping),
we define:

\begin{equation}\label{eq:option_space}
\renewcommand{\arraystretch}{1.2}
U'(x)=
{\small
\begin{cases}
\{u \in U(x)\mid f(x,u)\notin X_u\}, & \text{if } X_u \text{ given}, \\
\{u \in U(x)\mid f(x,u)\in X_d \cup X_o\}, & \text{if } X_d, X_o \text{ given}, \\
U(x), & \text{otherwise.}
\end{cases}
}
\end{equation}

Clients enforce safety by selecting decisions from $U'(x)$ whenever constraints are active.

\textbf{Signal 4: Action Prediction $u^\star \in U(x)$.}
An advisory recommendation for exploration-efficient information gathering, generated by the discovery service solely from client-agnostic telemetry. At information-gathering decision points, $u^\star$ identifies the query option most likely to yield high-priority information, enabling the client to bypass trial-and-error query selection. Clients may treat $u^\star$ as a heuristic suggestion, a priority ordering, or a direct selection depending on their autonomy design.

% figure
\begin{figure*}[t!]
	\centering
    \includegraphics[width=0.9\textwidth]{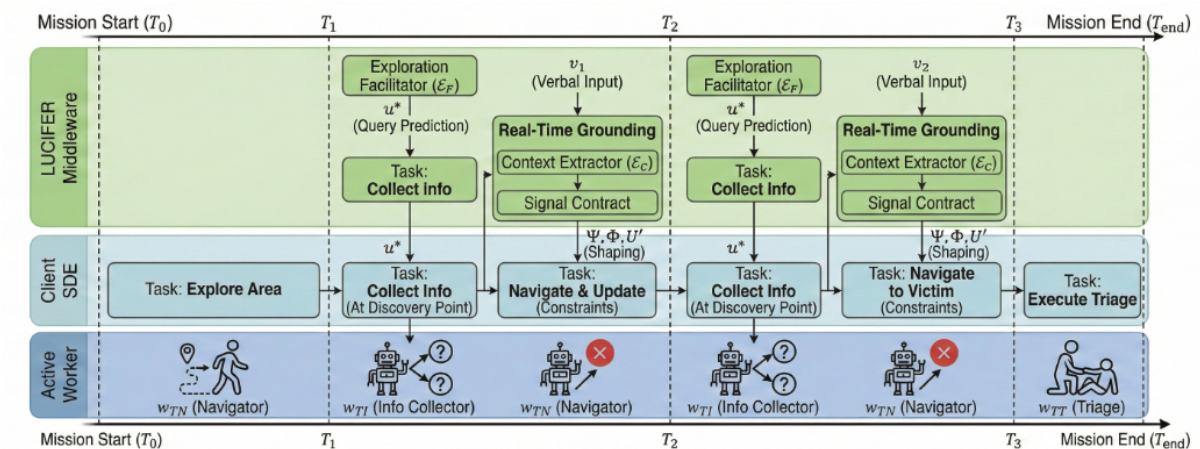}
	\caption{\textit{SAR Mission task Decomposition and LUCIFER Intervention:} The diagram illustrates the precise operational flow. Phase 1: \(w_{TN}\) navigates to discovery points (no LUCIFER). Phase 2: At discovery point, SDE activates \(w_{TI}\). Exploration Facilitator predicts the optimal query action \(u^*\). \(w_{TI}\) executes \(u^*\), receiving verbal input. Context Extractor immediately processes the input, generating shaping signals through the contract that refine subsequent navigation by \(w_{TN}\) (avoiding hazards and/or go through safe zones) and eventual triage by \(w_{TT}\).}
	\label{fig4:task_hierarhcy_timeline}
\end{figure*}

\section{Experimental Setup}
\label{sec:experimental_setup}
We use a dual-validation strategy to separate middleware reliability from client-specific decision-making. First, we benchmark the two middleware services, grounding ($\mathcal{E}_C$) and discovery ($\mathcal{E}_F$), as standalone components (\textit{infrastructure-level validation}). Second, we evaluate contract-level utility via controlled ablations with two structurally different clients that consume the \emph{same} Signal Contract (\textit{system-level validation}).

\textbf{Main system-level setting.}
All system-level results in the main text use the 3-info mission in a \textbf{$5\times5$} gridworld, evaluated across multiple layouts shared by both clients. We focus on whether the pattern is consistent across clients, that is grounding improves safety, discovery improves information-collection efficiency, and the combination yields both.

\subsection{Infrastructure Validation: Component Benchmarks}
\label{subsec:infra_validation}
We validate grounding and discovery independently before integration.

\subsubsection{Context Extractor Evaluation}
We compare LLM backends and traditional NLP~\cite{akbik2019flair} baselines (Fig.~\ref{fig3:verbal}) on three datasets (Standard, Messy, Advanced Messy) using Adjusted Accuracy (Eq.~\ref{eq:adjusted_accuracy}), isolating whether $\mathcal{E}_C$ reliably produces structured semantic objects $C$ under increasingly challenging linguistic phenomena. Rather than proposing a new taxonomy, we operationalise a stress-test protocol: Messy inputs include disfluencies and self-repairs (retract-and-replace patterns) consistent with classic accounts of disfluency/self-repair~\cite{1571698600491774336, hough2013modelling}, while Advanced Messy inputs require resolving implicit relational references (e.g., ``next to/attached to'') consistent with bridging-style inference settings studied in bridging resolution~\cite{hou-etal-2018-unrestricted}.

\begin{equation}
\text{Adj. Acc.} = \frac{\max(0, \text{Correct} - 2 \times \text{Hallucinations})}{\text{Total}} \times 100
\label{eq:adjusted_accuracy}
\end{equation}
where $\text{Correct} = (\text{Locations Found} + \text{Classifications Correct})$
and $\text{Total} = 2 \times \text{Expected Locations}$.

\subsubsection{Exploration Facilitator Evaluation}
We evaluate candidate LLMs as telemetry-only discovery backends, measuring prediction accuracy and latency for advisory query recommendations produced from learning client trace summaries.

\subsection{System Validation: Client-Agnostic Integration}
\label{subsec:system_validation}
We validate contract-level utility by holding each client fixed and varying only which contract signals are enabled.

\paragraph{Middleware ablations.}
For each client, we test four configurations:
\textbf{Baseline} (no middleware),
\textbf{Client+G} (grounding only),
\textbf{Client+D} (discovery only),
\textbf{Client+D+G} (both services).
We refer to the grounding configuration uniformly as \textbf{+G}. Grounding signals are incorporated natively by each client: as policy shaping for the RL client, and as constraint filtering for the planner.

\subsubsection{Environment and Task}
\label{subsubsec:env_task}
We simulate a SAR-inspired mission in a $5\times5$ non-sparse gridworld with multiple layouts. Each episode instantiates one layout from a fixed set to ensure that both clients operate under identical environment families.

\textbf{Mission objective (3-info).}
Fig.~\ref{fig4:task_hierarhcy_timeline} shows the operation flow. The agent must collect three required intelligence types (\emph{victim}, \emph{hazard}, \emph{safe route}) from information locations, then complete the mission objective (reach/finish at the designated terminal condition).

\textbf{Actions and interaction budget.}
The agent navigates with movement actions on the grid. At information locations, the agent can additionally execute 15 query actions (the information-collection action set). These query actions are the target of discovery ($\mathcal{E}_F$), which recommends (when active) the most promising query at the current information location.

\subsubsection{Client 1: Hierarchical RL Client}
\label{subsubsec:rl_client}
We use a hierarchical tabular RL client with a Strategic Decision Engine (SDE) that sequences navigation, information collection, and task completion workers \cite{panagopoulos2024selective}.

\textbf{Episode limits.}
Each episode is capped at 30 total environment steps. At information locations, the agent may spend up to 15 steps executing query actions (attempts) within the episode budget.

\textbf{Training and reporting protocol.}
RL variants are trained for 1000 episodes and results are aggregated over 40 independent runs (different random seeds/layout initializations, as applicable). Reported system-level metrics are the mean and standard deviation across runs.

\subsubsection{Client 2: Hybrid Planning Client}
\label{subsubsec:planner_client}
To test client-agnosticism, we implement a non-learning hybrid client that uses A$^\star$ for navigation and executes information collection decisions at information locations.

\textbf{Episode limits.}
Each run is capped at 100 total steps. At information locations, the agent may spend up to 15 steps executing query actions.

\textbf{No training.}
The planner is evaluated directly (no learning). Results are aggregated over 40 runs under the same environment family and layout set used for RL.

\textbf{Signal integration.}
When grounding is enabled, the planner enforces constraints $U'$ by filtering A$^\star$ neighbor generation to exclude hazard-adjacent transitions. When discovery is enabled, it directly executes $u^\star$ instead of random query selection.

\subsubsection{Performance Metrics}
\label{subsubsec:metrics}
We report three core metrics for both clients:

\begin{itemize}
    \item \textbf{Mission Success Rate (MSR)}: percentage of runs where the agent completes the full mission (collect all required information and satisfy the terminal objective).
    \item \textbf{Collection Success Rate (CSR)}: percentage of runs where the agent makes \emph{successful and correct} information-collection decisions. This metric isolates the \emph{quality of exploration decisions}, acting as a direct analogue to replacing random sampling in epsilon-greedy exploration.
    \item \textbf{Safe Mission Success (SMS)}: percentage of runs where the mission is completed \emph{and} the agent avoids all hazard collisions (safety outcome attributable to grounded constraints).
\end{itemize}

\subsubsection{Ablation Hypotheses}
\label{subsubsec:hypotheses}
We test necessity and synergy while keeping each client fixed:

\begin{itemize}
    \item \textbf{H1 (Grounding $\rightarrow$ Safety):} Enabling grounding (+G) enforces linguistically-derived constraints. Grounding establishes operational safety. Consequently, when active, SMS achieves parity with MSR, proving the middleware successfully prevents hazard collisions during completed runs.
    \item \textbf{H2 (Discovery $\rightarrow$ Collection Efficiency):} Enabling discovery (+D) improves exploration decision quality (CSR) without requiring discovery-specific training of the client, as LLM reasoning over structured trace summaries allows zero-shot generalisation to novel location signatures. However, it does not guarantee safety (SMS remains low without grounding).
    \item \textbf{H3 (Synergy):} Only +D+G achieves both safety (high SMS) and efficiency (high CSR), yielding the best overall mission performance (MSR).
    \item \textbf{H4 (Client-agnosticism):} The directional pattern (grounding$\rightarrow$safety, discovery$\rightarrow$efficiency, combined$\rightarrow$both) holds for \emph{both} clients, showing the effect is architectural rather than client-specific.
\end{itemize}

\begin{table}[t]
\centering
\caption{Context Extractor Performance Across LLM Models.}
\label{table1:informationLLMcomparison}
\footnotesize
\resizebox{\columnwidth}{!}{%
\begin{tabular}{lcccc}
\toprule
\multirow{2}{*}{\textbf{Model}} & \multicolumn{3}{c}{\textbf{Adjusted Accuracy (\%)}} & \multirow{2}{*}{\textbf{Avg Time (s)}} \\
\cmidrule(lr){2-4}
& \textbf{Standard} & \textbf{Messy} & \textbf{Adv. Messy} & \\
\midrule
Gemma2 (9B)      & 88.6 & 100.0 & 54.5 & 1.23  \\
Qwen3 (8B)   & 97.0 & 100.0 & 91.0 & 10.45 \\
Gemma3 (27B)      & 94.2 & 94.0  & 64.0 & 3.43  \\
GPT-OSS (7B)     & 98.3 & 99.3  & 77.7 & 4.32  \\
Qwen3 (32B)  & 99.4 & 100.0 & 81.3 & 22.52 \\
\bottomrule
\end{tabular}
} % end resizebox
\end{table}

\begin{table}[h]
\centering
\caption{Context Extractor vs. Traditional NLP Baselines.}
\label{table2:baseline_comparison}
\begin{tabular}{lccc}
\toprule
\textbf{Method} & \multicolumn{3}{c}{\textbf{Adjusted Accuracy (\%)}} \\
\cmidrule(lr){2-4}
& \textbf{Standard} & \textbf{Messy} & \textbf{Adv. Messy} \\
\midrule
Enhanced Rule-Based  & 75.3 & 20.8 & 22.5 \\
\quad (Traditional)  &      &      &      \\[0.5ex]
FLAIR NER            & 91.7 & 72.0 & 36.7 \\
\quad (Neural + Keywords) &  &      &      \\[0.5ex]
LLM (Qwen3 8B)       & 97.0 & \textbf{100.0} & 91.0 \\
\quad (Zero-Shot Reasoning) & & &   \\[1ex]
\midrule
\textit{Gap to LLM}  & \textit{+5.3} & \textit{+28.0} & \textit{+54.3} \\
\bottomrule
\end{tabular}
\end{table}

\section{Results and Analysis}
\label{sec:results}

We validate \textsc{LUCIFER} as middleware by answering three questions aligned with the architectural claim:
\textbf{(1) Grounding reliability:} can the Context Extractor $\mathcal{E}_C$ robustly translate messy reports into structured semantics?
\textbf{(2) Discovery reliability:} can the Exploration Facilitator $\mathcal{E}_F$ recommend high-value query actions using telemetry alone?
\textbf{(3) Contract-level utility:} when heterogeneous clients consume the same Signal Contract, do we observe the same directional pattern---grounding $\rightarrow$ safety, discovery $\rightarrow$ efficiency, and synergy when combined?

\subsection{Grounding Reliability: Context Extractor Robustness}
\label{sec:context_extractor_results}
Tables~\ref{table1:informationLLMcomparison}--\ref{table2:baseline_comparison} summarize grounding robustness.

\paragraph{LLMs remain robust under messy, self-correcting reports.}
Across models, adjusted accuracy is high on Standard inputs and remains strong on Messy inputs, indicating that pretrained LLMs can correctly resolve disfluencies and self-corrections that commonly break surface-form parsers.

\paragraph{Reasoning is necessary under implied references.}
Traditional baselines degrade sharply on Messy/Advanced Messy regimes, where the task requires selecting the speaker’s final intent and resolving implied spatial references. In contrast, $\mathcal{E}_C$ maintains high adjusted accuracy, supporting its use as an online grounding service that produces reliable structured objects $C$ for the Signal Contract.

\paragraph{Takeaway.}
These results validate that grounding can be treated as a standalone, inference-only middleware service whose errors are separable from downstream control errors.

\subsection{Discovery Reliability: Exploration Facilitator Backend}
\label{sec:exploration_facilitator_results}
Table~\ref{table3:LLM-predictor-comparison} reports the Exploration Facilitator’s backend reliability (accuracy and latency) under the same discovery calls issued during the RL client runs when discovery is enabled (+D). Concretely, the 99.8\% accuracy reported for the RL client under +D in Table~\ref{table:cross_client_comparison_sparse} corresponds to using \texttt{llama~3.1 (8B)} as the facilitator backend; \texttt{gemma3} is included for comparison.

\paragraph{Takeaway \& latency.}
We use Table~\ref{table3:LLM-predictor-comparison} to select llama~3.1 as the default discovery backend for system-level +D and +D+G evaluations. Its sub-second mean latency (0.7\,s) indicates that decoupled inference can meet interactive timing requirements in our testbed.

\begin{table}[t]
\begin{center}
% Updated caption to reflect specific setting
\caption{Performance Comparison of LLM Models as Exploration Facilitators (3-info)}
\label{table3:LLM-predictor-comparison}
\footnotesize
\setlength{\tabcolsep}{10pt} % Increased spacing slightly for readability
\renewcommand{\arraystretch}{0.95}
% Removed resizebox as the table is now compact enough to fit naturally
\begin{tabular}{l cc}
\toprule
\textbf{Model} & \textbf{Acc. (\%)} & \textbf{Time (s)} \\
\midrule
gemma3 (27B)    & 95.6\(\pm\)5.4 & 3.5\(\pm\)0.3 \\
llama3.1 (8B)  & 99.8\(\pm\)0.03 & 0.7\(\pm\)0.05 \\
\bottomrule
\end{tabular}
\end{center}
\end{table}

\subsection{System-Level Validation: Contract Necessity and Synergy (3-info)}
\label{sec:client_performance}
We test contract-level utility with two structurally different clients while varying only which contract signals are enabled (Table~\ref{table:cross_client_comparison_sparse}).

\paragraph{Understanding client baselines.}
The hybrid planner executes a single 100-step episode deterministically, whereas the RL client learns over 1000 episodes under exploration. This naturally yields different absolute baseline MSR/CSR values. Our claim concerns the direction and consistency of middleware effects under fixed client designs.

\paragraph{Grounding is necessary for safety (H1).}
Enabling grounding (+G) increases SMS across both clients. Because grounding establishes operational safety by filtering out hazard transitions, any successful mission \emph{must} be collision-free. Consequently, SMS matches MSR when +G is active, indicating that successful missions are collision-free by construction under enforced constraints. Grounding primarily shifts safety outcomes without guaranteeing efficient information collection.

\paragraph{Discovery is necessary for efficient information collection (H2).}
Enabling discovery (+D) improves exploration decision quality (higher CSR). By replacing random sampling with zero-shot LLM reasoning, +D enables efficient information collection. However, SMS remains low without grounding, confirming that discovery alone does not enforce safety.

\paragraph{Only the combined middleware achieves both (H3-H4).}
The combined (+D+G) configuration yields the strongest joint performance with improved SMS alongside improved CSR, while maintaining or improving MSR. Crucially, this directional pattern, grounding\(\rightarrow\)safety, discovery\(\rightarrow\)efficiency, synergy under combination, holds consistently for both learning and non-learning clients. This supports the claim that the effect is architectural rather than tied to a specific optimisation mechanism.

\paragraph{Takeaway.}
Across both clients, +G yields safety, +D yields efficiency, and +D+G yields their combination, supporting the contract hypothesis.

\begin{table}[t]
\centering
\footnotesize
\setlength{\tabcolsep}{2pt}
\renewcommand{\arraystretch}{1.2}
\caption{Cross-Client Performance in the 3-Info 5\(\times\)5 Setting. Metrics: MSR (Mission Success Rate), CSR (Collection Success Rate), SMS (Safe Mission Success). G = Grounding, D = Discovery. Grounding (+G) is instantiated natively by each client: as policy shaping for RL, and constraint filtering for the hybrid planner.}
\label{table:cross_client_comparison_sparse}

\begin{tabular}{@{}llccccc@{}}
\toprule
\multirow{2}{*}{\textbf{Client}} & \multirow{2}{*}{\textbf{Config}} & \multicolumn{3}{c}{\textbf{Performance Metrics (\%)}} & \textbf{Safety} & \textbf{Efficiency} \\
\cmidrule(lr){3-5}
 & & \textbf{MSR} & \textbf{CSR} & \textbf{SMS} & & \\
\midrule
\multirow{4}{*}{\shortstack[l]{RL\\(HierQ)}} 
 & Baseline & 49.2$\pm$8.4 & 3.9$\pm$0.8 & 16.5$\pm$20.9 & \ding{55} & \ding{55} \\
 & +G & 61.9$\pm$14.1 & 3.8$\pm$0.9 & 61.9$\pm$14.1 & \(\checkmark\) & \ding{55} \\
 & +D & 50.3$\pm$8.4 & 99.8$\pm$0.03 & 16.8$\pm$21.1 & \ding{55} & \(\checkmark\) \\
 & +D+G & 66.5$\pm$14.6 & 99.9$\pm$0.04 & 66.5$\pm$14.6 & \(\checkmark\) & \(\checkmark\) \\
\midrule
\multirow{4}{*}{\shortstack[l]{Hybrid\\(A*+Heur.)}} 
 & Baseline & 15.0$\pm$35.7 & 4.7$\pm$4.6 & 2.5$\pm$15.6 & \ding{55} & \ding{55} \\
 & +G & 25.0$\pm$43.3 & 4.8$\pm$5.6 & 25.0$\pm$43.3 & \(\checkmark\) & \ding{55} \\
 & +D & 100.0$\pm$0.0 & 92.0$\pm$16.0 & 15.0$\pm$35.7 & \ding{55} & \(\checkmark\) \\
 & +D+G & 100.0$\pm$0.0 & 96.8$\pm$10.0 & 100.0$\pm$0.0 & \(\checkmark\) & \(\checkmark\) \\
\bottomrule
\end{tabular}
\end{table}

\section{Discussion \& Limitations}
\label{sec:discussion}

The evidence from Sections \ref{sec:context_extractor_results}--\ref{sec:client_performance} supports three conclusions: (i) grounding is sufficiently reliable as an inference-only service whose failures are separable from downstream control errors in our setting; (ii) discovery is reliable and low-latency enough to serve as an online advisory backend; and (iii) their contract-level signals yield consistent safety and efficiency effects across heterogeneous clients. The contribution is therefore architectural, not a new planner or learning rule. Language understanding remains outside the client and enters only through bounded contract signals. The system-level ablations show that these signals remain actionable for heterogeneous consumers.

\subsection{Interpretation}\label{sec:viB}
\paragraph{What ``safety'' and ``efficiency'' mean here.}
Safety is operationalised as completing the mission without hazard collisions (SMS), capturing the practical effect of grounded constraints. Efficiency is operationalised by information-collection decision quality (CSR), which acts as a zero-shot reasoning replacement for random sampling during exploration, capturing whether the client selects high-value queries without wasting its interaction budget.

\paragraph{Why absolute performance differs across clients.}
Differences in absolute baseline MSR and CSR between the RL agent and the hybrid planner are expected given their different operating regimes (learning under exploration vs.\ deterministic replanning) and different step budgets. This does not affect the core claim, which concerns the \emph{direction} and \emph{consistency} of middleware effects under fixed client designs.

\subsection{Strengths}\label{sec:viC}
\paragraph{A stable, training-decoupled interface.}
By constraining middleware to inference-only processing and client-agnostic telemetry, the contract localises language updates to middleware configuration and isolates grounding faults from control faults for clearer diagnosis.

\paragraph{Client-agnostic validation.}
Demonstrating the same necessity/synergy pattern with two structurally different clients strengthens the case that the contract generalises across consumption mechanisms rather than exploiting a single client-specific integration trick.

\paragraph{Separable validation of reliability and utility.}
Component benchmarks establish that grounding and discovery are viable services on their own. System-level ablations then attribute downstream gains to enabling contract outputs rather than to client retuning.

\subsection{Limitations}\label{sec:viD}
\paragraph{Knowledge-base dependence.}
Grounding quality depends on the fidelity and freshness of the knowledge base $\mathcal{B}$ and the off-the-shelf LLMs. Retrieval mismatches can yield structurally valid but incorrect grounded objects, which may propagate into constraints or shaping signals.

\paragraph{No explicit uncertainty exposure for discovery.}
The discovery service currently provides a single recommendation without calibrated confidence. In ambiguous contexts, a confidence signal would enable principled gating (e.g., fall back to default query selection when uncertainty is high). Similarly to grounding, quality depends on LLM reasoning ability.

\paragraph{Assurance pathway enabled by the signal contract.}
Although we do not claim formal safety guarantees in this paper, the contract boundary provides a natural anchor for assurance-style monitoring. The client consumes only bounded mathematical objects (priors, potentials, admissible-action sets, and recommendations), and does not ingest raw language. This separation enables a compositional view in which (i) middleware enforces structural invariants by construction (e.g., bounded ranges, valid action sets) and (ii) runtime gates can be defined on top of exposed confidence/telemetry (e.g., activate constraints or shaping only when information is sufficiently supported, otherwise revert to a conservative fallback). Developing a full assurance argument that links such gates to downstream control properties is an important future direction.

\paragraph{Temporal evolution and contradiction handling.}
Our instantiation applies grounded updates at ingestion time but does not maintain an explicit temporal belief state (revision, decay, contradiction resolution). Longer deployments would benefit from belief maintenance policies orthogonal to the contract definition.

\paragraph{Trust and adversarial reporting.}
We evaluate robustness to unintentional messiness, not intentional deception or coordinated misinformation. Safety-critical deployment would require provenance and cross-validation mechanisms to mitigate malicious or erroneous reports.

\paragraph{Testbed abstraction and external validity.}
The $5\times5$ gridworld enables controlled attribution, but it abstracts real-world perception, continuous dynamics, and multi-agent coordination. Additional validation in higher-fidelity simulators and robotic stacks is needed to fully assess external validity.

\subsection{Practical Next Steps}\label{sec:viE}
Two extensions are immediately actionable without changing the contract: (i) expose confidence/uncertainty for discovery recommendations and (ii) add belief maintenance for evolving grounded knowledge (revision and conflict resolution). More broadly, evaluating the same contract in higher-fidelity SAR simulators and with additional client classes (e.g., rule-based safety monitors, MPC, model-based RL) would strengthen deployment relevance.

\section{Conclusion}
We presented \textsc{LUCIFER}, a training-decoupled middleware that converts deployment-time verbal reports into a standard Signal Contract consumed by heterogeneous downstream decision-makers. The middleware comprises (i) a grounding service ($\mathcal{E}_C$) that produces policy priors, reward potentials, and admissible-option constraints from streaming language, and (ii) a discovery service ($\mathcal{E}_F$) that recommends information-gathering actions from client-agnostic telemetry. In a SAR-inspired testbed with two structurally distinct clients (hierarchical tabular RL and a hybrid A$^\star$-heuristic planner), we found that: (1) semantic reasoning is required for robust grounding under messy, self-correcting and implicit-reference reports where pattern-based baselines degrade; (2) grounding and discovery address complementary failure modes (safety vs.\ information-collection efficiency), and neither alone suffices; and (3) enabling both yields the intended safety--efficiency synergy, with the same directional pattern across both clients. Beyond this testbed, the core contribution is an architectural pattern for \emph{online modular grounding} that localises language updates to middleware configuration (e.g., $\mathcal{I},\mathcal{B}$) rather than client retraining. Future work will test this transfer hypothesis in higher-fidelity SAR simulators, extend the contract with uncertainty and belief maintenance for temporally evolving reports, and evaluate additional client classes (e.g., model-based RL and MPC).

\bibliography{thebibliography}

@IEEEtranBSTCTL{IEEEexample:BSTcontrol,
  CTLdash_repeated_names = "no",
}

@article{da2024survey,
  title={A Survey on Human in the Loop for Self-Adaptive Systems},
  author={da Silva Tavares, Geov{\'a} Junio and Rosa, Nelson Souto},
  journal={Journal of Universal Computer Science},
  volume={30},
  number={12},
  pages={1626},
  year={2024},
  publisher={Pensoft Publishers}
}

@inproceedings{kruijff2013experience,
  title={Experience in system design for human-robot teaming in urban search and rescue},
  author={Kruijff, Geert-Jan M and Jan{\'\i}{\v{c}}ek, M and Keshavdas, Shanker and Larochelle, Benoit and Zender, Hendrik and Smets, Nanja JJM and Mioch, Tina and Neerincx, Mark A and Diggelen, JV and Colas, Francis and others},
  booktitle={Field and Service Robotics: Results of the 8th International Conference},
  pages={111--125},
  year={2013},
  organization={Springer}
}

@article{tellex2020robots,
  title={Robots that use language},
  author={Tellex, Stefanie and Gopalan, Nakul and Kress-Gazit, Hadas and Matuszek, Cynthia},
  journal={Annual Review of Control, Robotics, and Autonomous Systems},
  volume={3},
  number={1},
  pages={25--55},
  year={2020},
  publisher={Annual Reviews}
}

@article{cohen2024survey,
  title={A survey of robotic language grounding: Tradeoffs between symbols and embeddings},
  author={Cohen, Vanya and Liu, Jason Xinyu and Mooney, Raymond and Tellex, Stefanie and Watkins, David},
  journal={arXiv preprint arXiv:2405.13245},
  year={2024}
}

@inproceedings{du2023guiding,
  title={Guiding pretraining in reinforcement learning with large language models},
  author={Du, Yuqing and Watkins, Olivia and Wang, Zihan and Colas, C{\'e}dric and Darrell, Trevor and Abbeel, Pieter and Gupta, Abhishek and Andreas, Jacob},
  booktitle={International Conference on Machine Learning},
  pages={8657--8677},
  year={2023},
  organization={PMLR}
}

@article{ahn2022can,
  title={Do as i can, not as i say: Grounding language in robotic affordances},
  author={Ahn, Michael and Brohan, Anthony and Brown, Noah and Chebotar, Yevgen and Cortes, Omar and David, Byron and Finn, Chelsea and Fu, Chuyuan and Gopalakrishnan, Keerthana and Hausman, Karol and others},
  journal={arXiv preprint arXiv:2204.01691},
  year={2022}
}

@misc{cheng2023llfbenchbenchmarkinteractivelearning,
  title={LLF-Bench: Benchmark for Interactive Learning from Language Feedback},
  author={Ching-An Cheng and Andrey Kolobov and Dipendra Misra and Allen Nie and Adith Swaminathan},
  year={2023},
  eprint={2312.06853},
  archivePrefix={arXiv},
  primaryClass={cs.AI},
  url={https://arxiv.org/abs/2312.06853},
}

@article{pirinen2022aerial,
  title={Aerial view localization with reinforcement learning: Towards emulating search-and-rescue},
  author={Pirinen, Aleksis and Samuelsson, Anton and Backsund, John and {\AA}str{\"o}m, Kalle},
  journal={arXiv preprint arXiv:2209.03694},
  year={2022}
}

@inproceedings{gruffeille2024disaster,
  title={Disaster area coverage optimisation using reinforcement learning},
  author={Gruffeille, Ciaran and Perrusqu{\'\i}a, Adolfo and Tsourdos, Antonios and Guo, Weisi},
  booktitle={2024 International Conference on Unmanned Aircraft Systems (ICUAS)},
  pages={61--67},
  year={2024},
  organization={IEEE}
}

@inproceedings{murphy2000mobility,
  title={Mobility and sensing demands in USAR},
  author={Murphy, Robin and Casper, Jennifer and Hyams, Jeffery and Micire, Mark and Minten, Brian},
  booktitle={2000 26th Annual Conference of the IEEE Industrial Electronics Society. IECON 2000. 2000 IEEE International Conference on Industrial Electronics, Control and Instrumentation. 21st Century Technologies},
  volume={1},
  pages={138--142},
  year={2000},
  organization={IEEE}
}

@inproceedings{saeed2019role,
  title={Role of stakeholders in mitigating disaster prevalence: Theoretical Perspective},
  author={Saeed, AL-Fazari and Kasim, Narimah},
  booktitle={MATEC Web of Conferences},
  volume={266},
  pages={03008},
  year={2019},
  organization={EDP Sciences}
}

@article{ter2023stakeholder,
  title={Stakeholder Analysis in the Context of Natural Disaster Mitigation: The Case of Flooding in Three US Cities},
  author={Ter-Mkrtchyan, Ani V and Franklin, Aimee L},
  journal={Sustainability},
  volume={15},
  number={20},
  pages={14945},
  year={2023},
  publisher={MDPI}
}

@misc{luketina2019surveyreinforcementlearninginformed,
  title={A Survey of Reinforcement Learning Informed by Natural Language},
  author={Jelena Luketina and Nantas Nardelli and Gregory Farquhar and Jakob Foerster and Jacob Andreas and Edward Grefenstette and Shimon Whiteson and Tim Rockt{\"a}schel},
  year={2019},
  eprint={1906.03926},
  archivePrefix={arXiv},
  primaryClass={cs.LG},
  url={https://arxiv.org/abs/1906.03926},
}

@article{xie2023text2reward,
  title={Text2reward: Reward shaping with language models for reinforcement learning},
  author={Xie, Tianbao and Zhao, Siheng and Wu, Chen Henry and Liu, Yitao and Luo, Qian and Zhong, Victor and Yang, Yanchao and Yu, Tao},
  journal={arXiv preprint arXiv:2309.11489},
  year={2023}
}

@article{ma2023eureka,
  title={Eureka: Human-level reward design via coding large language models},
  author={Ma, Yecheng Jason and Liang, William and Wang, Guanzhi and Huang, De-An and Bastani, Osbert and Jayaraman, Dinesh and Zhu, Yuke and Fan, Linxi and Anandkumar, Anima},
  journal={arXiv preprint arXiv:2310.12931},
  year={2023}
}

@article{song2023llm,
  title={Llm-planner: Few-shot grounded planning for embodied agents with large language models},
  author={Song, Chan Hee and Wu, Jiaman and Washington, Clayton and Sadler, Brian M and Chao, Wei-Lun and Su, Yu},
  booktitle={Proceedings of the IEEE/CVF International Conference on Computer Vision},
  pages={2998--3009},
  year={2023}
}

@inproceedings{shah2023navigation,
  title={Navigation with large language models: Semantic guesswork as a heuristic for planning},
  author={Shah, Dhruv and Equi, Michael Robert and Osi{\'n}ski, B{\l}a{\.z}ej and Xia, Fei and Ichter, Brian and Levine, Sergey},
  booktitle={Conference on Robot Learning},
  pages={2683--2699},
  year={2023},
  organization={PMLR}
}

@article{lynch2023interactive,
  title={Interactive language: Talking to robots in real time},
  author={Lynch, Corey and Wahid, Ayzaan and Tompson, Jonathan and Ding, Tianli and Betker, James and Baruch, Robert and Armstrong, Travis and Florence, Pete},
  journal={IEEE Robotics and Automation Letters},
  year={2023},
  publisher={IEEE}
}

@article{liu2022instruction,
  title={Instruction-following agents with multimodal transformer},
  author={Liu, Hao and Lee, Lisa and Lee, Kimin and Abbeel, Pieter},
  journal={arXiv preprint arXiv:2210.13431},
  year={2022}
}

@inproceedings{karamcheti2022lila,
  title={Lila: Language-informed latent actions},
  author={Karamcheti, Siddharth and Srivastava, Megha and Liang, Percy and Sadigh, Dorsa},
  booktitle={Conference on robot learning},
  pages={1379--1390},
  year={2022},
  organization={PMLR}
}

@inproceedings{colas2023augmenting,
  title={Augmenting autotelic agents with large language models},
  author={Colas, C{\'e}dric and Teodorescu, Laetitia and Oudeyer, Pierre-Yves and Yuan, Xingdi and C{\^o}t{\'e}, Marc-Alexandre},
  booktitle={Conference on Lifelong Learning Agents},
  pages={205--226},
  year={2023},
  organization={PMLR}
}

@article{rocamonde2023vision,
  title={Vision-language models are zero-shot reward models for reinforcement learning},
  author={Rocamonde, Juan and Montesinos, Victoriano and Nava, Elvis and Perez, Ethan and Lindner, David},
  journal={arXiv preprint arXiv:2310.12921},
  year={2023}
}

@article{rana2023sayplan,
  title={Sayplan: Grounding large language models using 3d scene graphs for scalable robot task planning},
  author={Rana, Krishan and Haviland, Jesse and Garg, Sourav and Abou-Chakra, Jad and Reid, Ian and Suenderhauf, Niko},
  journal={arXiv preprint arXiv:2307.06135},
  year={2023}
}

@article{qi2024safety,
  title={Safety control of service robots with LLMs and embodied knowledge graphs},
  author={Qi, Yong and Kyebambo, Gabriel and Xie, Siyuan and Shen, Wei and Wang, Shenghui and Xie, Bitao and He, Bin and Wang, Zhipeng and Jiang, Shuo},
  journal={arXiv preprint arXiv:2405.17846},
  year={2024}
}

@article{lewis2020retrieval,
  title={Retrieval-augmented generation for knowledge-intensive nlp tasks},
  author={Lewis, Patrick and Perez, Ethan and Piktus, Aleksandra and Petroni, Fabio and Karpukhin, Vladimir and Goyal, Naman and K{\"u}ttler, Heinrich and Lewis, Mike and Yih, Wen-tau and Rockt{\"a}schel, Tim and others},
  journal={Advances in neural information processing systems},
  volume={33},
  pages={9459--9474},
  year={2020}
}

@inproceedings{panagopoulos2024selective,
  title={Selective exploration and information gathering in search and rescue using hierarchical learning guided by natural language input},
  author={Panagopoulos, Dimitrios and Perrusquia, Adolfo and Guo, Weisi},
  booktitle={2024 IEEE International Conference on Systems, Man, and Cybernetics (SMC)},
  pages={1175--1180},
  year={2024},
  organization={IEEE}
}

@inproceedings{akbik2019flair,
  title={FLAIR: An easy-to-use framework for state-of-the-art NLP},
  author={Akbik, Alan and Bergmann, Tanja and Blythe, Duncan and Rasul, Kashif and Schweter, Stefan and Vollgraf, Roland},
  booktitle={Proceedings of the 2019 conference of the North American chapter of the association for computational linguistics (demonstrations)},
  pages={54--59},
  year={2019}
}

@article{hou-etal-2018-unrestricted,
  title = "Unrestricted Bridging Resolution",
  author = "Hou, Yufang  and Markert, Katja  and Strube, Michael",
  journal = "Computational Linguistics",
  volume = "44",
  number = "2",
  month = jun,
  year = "2018",
  address = "Cambridge, MA",
  publisher = "MIT Press",
  url = "https://aclanthology.org/J18-2002/",
  doi = "10.1162/COLI_a_00315",
  pages = "237--284"
}

@article{1571698600491774336,
  author="SHRIBERG E. E.",
  title="Preliminaries to a theory of speech disfluencies",
  journal="Doctoral dissertation, University of California at Berkeley",
  year="1994",
  URL="https://cir.nii.ac.jp/crid/1571698600491774336"
}

@article{hough2013modelling,
  title={Modelling expectation in the self-repair processing of annotat-, um, listeners},
  author={Hough, Julian and Purver, Matthew and others},
  year={2013},
  publisher={University of Amsterdam}
}

@inproceedings{tarakli2024interactive,
  title={Interactive reinforcement learning from natural language feedback},
  author={Tarakli, Imene and Vinanzi, Samuele and Di Nuovo, Alessandro},
  booktitle={2024 IEEE/RSJ International Conference on Intelligent Robots and Systems (IROS)},
  pages={11478--11484},
  year={2024},
  organization={IEEE}
}
\bibliographystyle{IEEEtran}

\end{document}